# MAINTENANCE IN PROBABILISTIC KNOWLEDGE-BASED SYSTEMS

Thomas F. Reid
Gregory S. Parnell
Bruce W. Morlan

Department of Operational Sciences
School of Engineering
Air Force Institute of Technology
Wright-Patterson AFB OH  45433

## ABSTRACT

Recent developments using directed acyclical graphs (i.e., influence diagrams and Bayesian networks) for knowledge representation have lessened the problems of using probability in knowledge-based systems (KBS). Most current research involves the efficient propagation of new evidence, but little has been done concerning the maintenance of domain-specific knowledge, which includes the probabilistic information about the problem domain. By making use of conditional independencies represented in the graphs, however, probability assessments are required only for certain variables when the knowledge base is updated.

The purpose of this study was to investigate, for those variables which require probability assessments, ways to reduce the amount of new knowledge required from the expert when updating probabilistic information in a probabilistic knowledge-based system. Three special cases (ignored outcome, split outcome, and assumed constant outcome) were identified under which many of the original probabilities (those already in the knowledge-base) do not need to be reassessed when maintenance is required.

## *Introduction*

Just as with rule-based systems, the general domain knowledge of probabilistic knowledge-based systems is not static. The KBS must be changed to reflect the changed state of information when new tests for existing hypotheses are developed, new hypotheses are formed, or a more thorough understanding of the problem domain is gained. When determined that the KBS model represented in the influence diagram is no longer adequate, the knowledge engineer's first task is the reassessment of the nodes and their dependencies. Nodes may be added or deleted, outcome spaces for individual variables may increase or decrease, arcs may be added or deleted, or the probability distributions for a variable's outcomes may be changed. Because of the local modularity property, the only probability distributions that must be re-encoded are those associated with nodes that have had some change made to their outcome space (gaining, losing, or changing outcomes) or incoming arcs (gaining or losing an incoming arc, or having the outcome space of a conditioning variable modified) (Heckerman & Horvitz:125).

In systems where the dependency structure changes infrequently, the time and effort required to encode the new distributions may be relatively insignificant. However, as these systems are applied to problem domains which are highly dynamic, i.e., the dependencies and probabilities often change, a significantly larger portion of time will be spent encoding probabilities. This paper examines ways in which the probabilistic information can change and possible means to reduce the effort required in the encoding process.

*What Happens When the State of Information Changes?*

When the state of information changes, the dependency structure for the influence diagram must be reassessed and those nodes which experience a change in their incoming arcs or outcome space must be reassessed. The more nodes that experience such changes, the more information that must be encoded from the expert. At the very least, probabilistic information must be encoded for the new outcomes and new variables. Also, any data invalidated by the change in the state of information must be reassessed, even if the dependency structure did not change. However, all is not necessarily lost. There may be some circumstances under which all, or nearly all, of the original probabilistic information is still valid. Some of these circumstances are identified in the following sections as special cases which may apply for some state of information changes.

*Special Cases for Marginal and Conditional Probability Distributions*

As indicated by Pearl, Shachter, and others, information from experts is more easily gathered in the form of marginal and conditional distributions (2:5; 3:246; 4:55). Since information is primarily collected in this manner, it makes sense to examine possible effort-saving special cases from this perspective. The primary objective is to keep as many of the original probabilities as possible relevant under the new state of information.

Special cases based on the marginal and conditional distributions can be readily grouped into those applicable when: 1) the outcome space for a variable changes in size; 2) a variable is added or removed from the influence diagram; and 3) an arc between two nodes is added or removed, changing the conditioning information in the diagram. The only other change which indicates a new state of information is when underlying probabilities change. No special cases were found to reduce the number of assessments required in response to this type of change.

For each special case, we examine, the effects on the node being changed (either a node experiencing a change in its outcome space, or a new, added node) and on nodes whose incoming arcs are somehow modified (either by a change in the outcome space of a conditional predecessor, or by the addition or loss of conditional predecessors). Since exponential growth can occur, these special cases were developed primarily with an expansion of the outcome space or number of variables in mind.

*Changes in the Outcome Space.* When the new state of information changes a node's outcome space, the probability distribution for that node must be reassessed. The distribution of any other nodes which were previously, or are now, conditioned on the changed node must also be reassessed. Two special cases, the "ignored outcome" and the "split outcome", may reduce the number of assessments required.

*Ignored Outcome Special Case.* Interest in the first special case was motivated by the following question: if a new, or previously "forgotten" outcome was added to a node, under what conditions would the original probabilistic information be of use? For this case, the original outcome space for the changed variable would be mutually exclusive but not collectively exhaustive.

Let $A$ be a node with m outcomes under the original state of information, $\mathcal{E}$. Now the expert perceives a previously ignored outcome, $a_{m+1}$. This new knowledge (that outcome $a_{m+1}$ exists) indicates a change in the state of information, and the diagram must be reassessed relative to this new state of information, $\mathcal{E}'$. If the expert determines that the old probability distribution for $A$, given the conditional predecessors $C(A)$, is

$$P[A=a_i \mid C(A), \mathcal{E}] = P[A=a_i \mid C(A), A \neq a_{m+1}, \mathcal{E}'] \qquad (1)$$

then the new probabilities for the original outcomes of $A$ are given by

$$P[A=a_i \mid C_j(A), \mathcal{E}'] = \lambda_j * P[A=a_i \mid C_j(A), \mathcal{E}] \quad \begin{aligned} i &= 1, \ldots, m \\ j &= 1, \ldots, \prod_{X \in C(A)} \|X\| \end{aligned} \qquad (2)$$

where $\|X\|$ denotes the number of outcomes for variable $X$ and $\lambda_j$ is a scaling factor for the probability distribution of $A$ given the old state of information and a specific combination, indexed by the subscript j, of the outcomes for variables in $C(A)$, and is given by

$$\lambda_j = 1 - P[A=a_{m+1} \mid C_j(A), \mathcal{E}'] \qquad (3)$$
$$= P[A \neq a_{m+1} \mid C_j(A), \mathcal{E}']$$

When considering the addition of $k$ new outcomes (instead of just one), the primary difference is in the calculation of the $\lambda_j$.

$$\lambda_j = 1 - \sum_{i=m+1}^{m+k} P[A=a_i \mid C_j(A), \mathcal{E}']$$

This means, for the expanded variable $A$, only conditional probabilities for the new outcomes must be encoded. Once these are obtained, a $\lambda_j$ for each combination of outcomes of $C(A)$ can be computed directly, and the probabilities under $\mathcal{E}'$ for the original outcomes are given by Eq (2).

The reduction in the required number of encodings depends on the number of old ($m$) and new ($k$) outcomes for $A$, the number of conditional predecessors for $A$ ($\|C(A)\|$), and the number of outcomes for each predecessor. For comparative purposes, suppose that each conditional predecessor of $A$ has $n$ outcomes[1]. Then the number of encodings needed to determine $A$'s distribution in the general case is $(m+k-1) \times n^{\|C(A)\|}$, since probabilities for all but one of the $A$'s $m+k$ outcomes are needed for each combination of the outcomes in $C(A)$. Similarly, the number of required probability assessments for the ignored outcome special case is just $k \times n^{\|C(A)\|}$, because only the probabilities for the new outcomes of $A$ are needed.

---

[1] This supposition is only made for notational convenience. All results remain valid when the number of outcomes is allowed to vary for each conditional predecessor, but the number of combinations of those outcomes is calculated differently.

A similar reduction can be found in the number of probability assessments for direct successors of nodes with increased outcome spaces. The applicability of the special case must be assessed for each direct successor node individually. When the probability distribution for $B$, a direct successor of $A$ under the new state of information $\mathcal{E}'$, is given by

$$P[B=b_j \mid A=a_i, C(B), \mathcal{E}'] = P[B=b_j \mid A=a_i, C(B), \mathcal{E}] \quad i=1,\ldots,m \quad (5)$$

then the only conditional distributions which must be assessed for $B$ are those which are conditioned on $a_{m+1}, \ldots, a_{m+k}$, the new outcomes of $A$. This means the original distributions for $B$ are still valid under the new state of information.

The number of probability assessments required to determine $B$'s distribution depends on $m$, $k$, the number of outcomes for $B$ ($p$), and the number of outcomes for each of the variables in $C(B)\backslash A^2$. For the general case, $(m+k)\times(p-1)\times n^{\|C(B)\backslash A\|}$ probability assessments are needed. This is reduced to $k\times(p-1)\times n^{\|C(B)\backslash A\|}$ when this special case applies.

Figures 1.a and 1.b show a graphical comparison of the relative number of assessments required in the general case and the number required in the ignored outcome case.

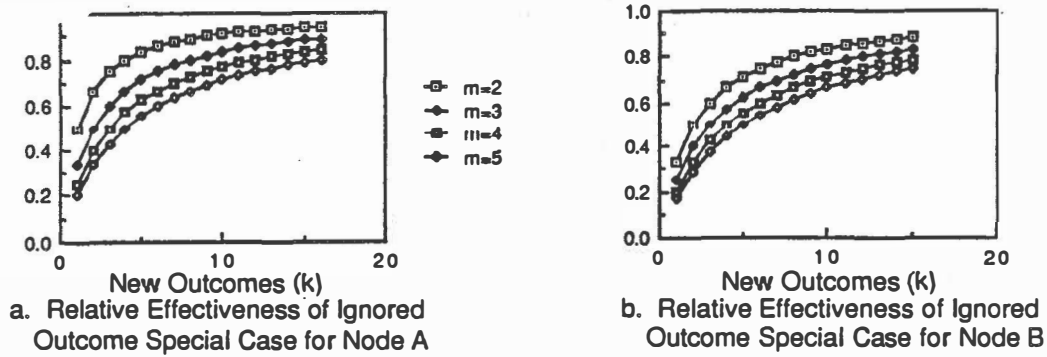

a. Relative Effectiveness of Ignored Outcome Special Case for Node A

b. Relative Effectiveness of Ignored Outcome Special Case for Node B

Figure 1. Ignored Outcome Versus General Case Data Requirements
When Expanding $A$ by $k$ Outcomes

For each indicated value of $m$, the curves show the ratio of the number of assessments required for the special case to the number required in the general case. This ratio is $\frac{k}{m+k-1}$ for Figure 1.a, and $\frac{k}{m+k}$ for Figure 1.b. If $A$ originally has two outcomes, then gains another due to a change in the state of information, Figure 1.a shows that the special case requires 1/2 as many probability assessments as the general case to determine the distribution for $A$, and Figure 1.b shows that only 1/3 as many are needed to determine the distribution for $B$.

---

$^2 C(B)\backslash A$ denotes the set of all conditional predecessors of $B$, excluding $A$.

Notice that as the number of new outcomes increases, the relative effectiveness of the special case decreases. Conversely, as $m$ increases, the relative effectiveness of the special case increases. Both of these reflect that the effectiveness of the special case depends on the amount of growth relative to the amount of data for the given distribution in the original influence diagram.

*Split Outcome Special Case.* A similar special case exists for situations where an outcome of a variable, say $A$, is split into two or more distinct outcomes. In effect, the original outcome, say $A = a_s$, was actually many outcomes: $a_{s1}, a_{s2}, \ldots, a_{sk}$. Unless the expert determines otherwise, the probabilities for the unchanged outcomes are still valid under the new state of information. That is,

$$P[A = a_i \mid C(A), \mathcal{E}''] = P[A = a_i \mid C(A), \mathcal{E}] \quad i \neq s \qquad (6)$$

for $a_i$ in the set of unchanged outcomes of $A$. The conditional probabilities for the new outcomes can then be assessed directly, where

$$\sum_{i=1}^{k} P[A = a_{si} \mid C(A), \mathcal{E}''] = P[A = a_s \mid C(A), \mathcal{E}] \qquad (7)$$

for each combination of outcomes of the conditional predecessors of $A$.

Just as for the ignored outcome case, direct successors of a "split outcome" node do not have to have probabilities reassessed which are dependent on the unchanged outcomes of $A$. Only probabilities conditioned on the new $a_{si}$ outcomes need be assessed. Thus the new distribution is given by

$$P[A = a_i \mid C(A), \mathcal{E}''] = P[A = a_i \mid C(A), \mathcal{E}] \quad i \neq s \quad \text{and} \qquad (8)$$

$$P[A = a_{si} \mid C_j(A), \mathcal{E}''] = \lambda_{ij} P[A = a_s \mid C_j(A), \mathcal{E}] \quad \begin{array}{l} i = 1, \ldots, k \\ j = 1, \ldots, \prod_{X \in C(A)} \|X\| \end{array} \qquad (9)$$

where $\sum_{i=1}^{k} \lambda_{ij}$ equals one for each combination $C_j(A)$ of $A$'s conditional predecessors.

Since $A$ now has $m + k - 1$ outcomes the number of assessments required (for the general case) is $(m + k - 2) \times n^{\|C(A)\|}$. When the split outcome case applies, this is reduced to $(k - 1) \times n^{\|C(A)\|}$.

Any direct successor $(B)$ of $A$ may not need all of its probabilities reassessed. If the the expert determines that the original conditional probabilities for $B$, given outcomes $a_{i, i \neq s}$ of $A$, are still valid, then only the probabilities concerning $A$'s new outcomes must be assessed. If $B$ has $p$ outcomes, and each variable in $C(B) \setminus A$ has $n$ outcomes, then for the general case $(m + k - 1) \times (p - 1) \times n^{\|C(B) \setminus A\|}$ probability assessments are required. If the split outcome case applies, the number of assessments is reduced to $k \times (p-1) \times n^{\|C(B)\setminus A\|}$.

The graphs in Figures 2.a and 2.b show the effectiveness of the split outcome special case relative to the general case. Similar to the graphs in Figure 1, these graphs show the ratio of the required number of assessments for the split outcome special case to those for the general case: $\frac{k-1}{m+k-2}$ for Figure 2.a, and $\frac{k}{m+k-1}$ for Figure 2.b. Again, notice the same type of effect from the relative size of the increase in $A$: as $k$ becomes large relative to $m$, the effectiveness of the special case decreases.

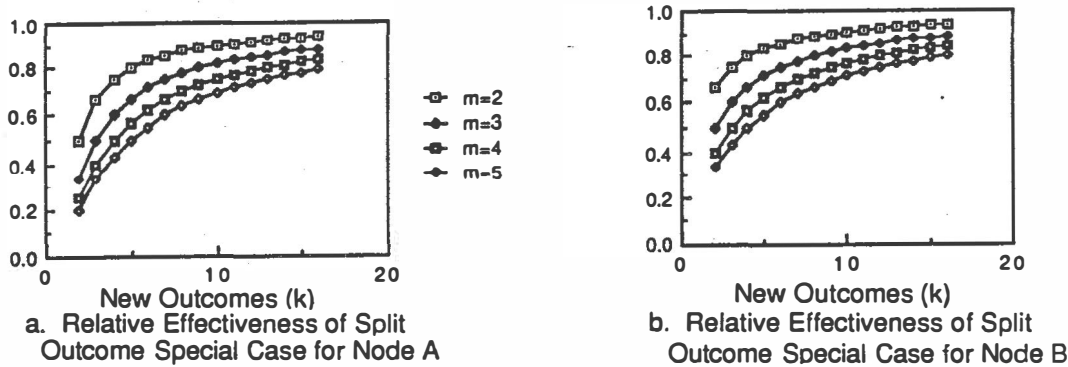

a. Relative Effectiveness of Split Outcome Special Case for Node A

b. Relative Effectiveness of Split Outcome Special Case for Node B

Figure 2. Split Outcome Versus General Case Data Requirements When Splitting an Outcome of $A$ Into $k$ Outcomes

If the variable $A$ is added to an influence diagram, the number of probability assessments required to define $A$'s distribution depends on the number of outcomes for $A$ and the number of outcomes for each conditional predecessor of $A$. Since there was previously no information in the diagram regarding $A$, all of these probabilities must be assessed. Additionally, the variables which now have $A$ as a conditional predecessor must now have their distributions reassessed. One special case, the "assumed constant outcome" case, was identified.

*Assumed Constant Outcome Special Case.* One way that the number of assessments can be reduced is if the old state of information, $\&$, is just the new state of information with the added condition that $A = a_0$. This might be the case when an expert learns that a factor previously considered constant did, in fact, have additional outcomes. Part of the probabilities, for nodes which gain $A$ as a conditional predecessor, would then transfer directly from the original state of information to the new state of information. Suppose $B$ is a node with $p$ outcomes that is conditioned on the newly added $A$. If this special case applies,

$$P[B=b_j \mid A=a_0, C(B), \&'] = P[B=b_j \mid C(B), \&] \quad j = 1, \ldots, p \qquad (10)$$

If each conditional predecessor (other than $A$) of $B$ has $n$ outcomes, the number of assessments drops from $k \times (p-1) \times n^{\|C(B)\setminus A\|}$ for the general case to $(k-1) \times (p-1) \times n^{\|C(B)\setminus A\|}$ for the special case.

The graphs in Figures 3.a and 3.b show the relative number of assessments required for the assumed constant outcome special case. Since this special case does not reduce the required number of assessments to determine the probability distribution of the newly added variable, Figure 3.a shows that exactly the same number of probabilities must be assessed (ratio = 1). Figure 3.b, like Figures 1 and 2, shows the decreasing effectiveness of this special case as k becomes larger for any nodes which gain $A$ as a conditional predecessor and for which the special case applies. The ratio for this graph is given by $\frac{k-1}{k}$.

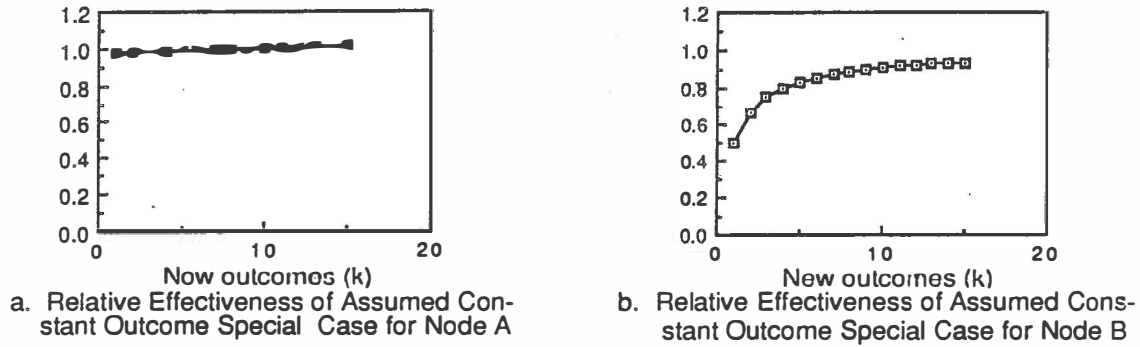

a. Relative Effectiveness of Assumed Constant Outcome Special Case for Node A

b. Relative Effectiveness of Assumed Constant Outcome Special Case for Node B

Figure 3. Assumed Constant Outcome Versus General Case Data Requirements When Adding New Variable $A$

*Changes in Conditioning.* When an arc is added between two nodes, say from $A$ to $B$, only $B$ must have its probability distribution reassessed. Since the distribution for the predecessor node ($A$) is defined by the conditional predecessors of $A$, no changes must be made to $A$'s distribution. The effect on $B$, however, is one seen earlier, in the discussion concerning changes in the number of variables. In fact, adding an arc can be viewed as a special case of adding a new variable. Thus the assumed constant outcome special case may also be applicable when adding a new arc between existing nodes. For $B$, the new direct successor of $A$, the situation is identical to that for a successor of a newly added node: if the original probability distribution of $B$ is valid for one outcome of $A$, those values need not be gathered again.

*The Importance of Conditional Independence.* The importance of getting the correct conditioning relationships in the influence diagram can not be overstated. If valid relationships are left out, the domain-specific knowledge base will be incomplete, and the system may reach conclusions that differ greatly from the expert's. This would eventually lead to a maintenance action to correct the discrepancy, much as rule-based systems are updated when they reach incorrect conclusions. If unnecessary arcs are included in the diagram, the number of probabilities which must be assessed is unduly increased. The conclusions will be the same as those reached using the diagram without the unnecessary arc, but more steps (i.e., more computer resources) will be required to reach those conclusions.

*Summary*

Three special cases (ignored outcome, split outcome, and assumed constant outcome) point to situations where part or all of the marginal and conditional probabilities for nodes with changed incoming arcs can be used under the new state of information. Although these special cases do provide some decrease in the number of probability assessments which must be done to complete the modified influence diagram, their applicability must be determined, by the expert, for each change that is made. Since the number of required assessments, even when a special case applies, is exponential in the number of conditional predecessors to the node being reassessed, the selection of the minimum essential conditioning relationships is much more important in keeping the number of assessments as low as possible.

*Bibliography*

1. Heckerman, David E. and Eric J. Horvitz. "On the Expressiveness of Rule-based Systems for Reasoning with Uncertainty," *Proceedings AAAI-87: Sixth National Conference on Artificial Intelligence.* 121-126. Los Altos CA: Morgan Kaufmann Publishers, Inc. 1987.

2. Henrion, Max. "Uncertainty in Artificial Intelligence: Is Probability Epistemologically and Heuristically Adequate?" To Appear in "Expert Systems and Expert Judgment," *Proceedings of the NATO Advanced Research Workshop.* Porto, Portugal, August, 1986.

3. Pearl, Judea. "Fusion, Propagation, and Structuring in Belief Networks," *Artificial Intelligence, 29:* 241-288 (September 1986).

4. Shachter, Ross D. and David E. Heckerman. "Thinking Backward for Knowledge Acquisition," *AI Magazine:* 55-61 (Fall 1987).